\documentclass[11pt,a4paper]{article}
\usepackage[hyperref]{acl2019}
\usepackage{times}
\usepackage{latexsym}
\usepackage{graphicx}
\usepackage{amsfonts}
\usepackage{amsmath}
\usepackage{multirow}
\usepackage{booktabs}
\usepackage{makecell}
\usepackage{url}
\usepackage{float}

\aclfinalcopy

\newcommand{\ignore}[1]{}

\title{Simple BERT Models for Relation Extraction\\ and Semantic Role Labeling}

\author{Peng Shi \and Jimmy Lin \vspace{0.1cm}\\
  David R. Cheriton School of Computer Science \\
  University of Waterloo\\
  \texttt{\{peng.shi, jimmylin\}@uwaterloo.ca} \\}

\date{}

\begin{document}
\maketitle
\begin{abstract}
We present simple BERT-based models for relation extraction and semantic role labeling.
In recent years, state-of-the-art performance has been achieved using neural models by incorporating lexical and syntactic features such as part-of-speech tags and dependency trees.
In this paper, extensive experiments on datasets for these two tasks show that without using any external features, a simple BERT-based model can achieve state-of-the-art performance.
To our knowledge, we are the first to successfully apply BERT in this manner.
Our models provide strong baselines for future research.
\end{abstract}

\section{Introduction}

Relation extraction and semantic role labeling~(SRL) are two fundamental tasks in natural language understanding.
The task of relation extraction is to discern whether a relation exists between two entities in a sentence. 
For example, in the sentence ``Obama was born in Honolulu'', ``Obama'' is the subject entity and ``Honolulu'' is the object entity. 
The task of a relation extraction model is to identify the relation between the entities, which is \texttt{\small per:city\_of\_birth} (birth city for a person). 
For SRL, the task is to extract the predicate--argument structure of a sentence, determining ``who did what to whom'', ``when'', ``where'', etc. 
Both capabilities are useful in several downstream tasks such as question answering~\cite{shen2007using} and open information extraction~\cite{fader2011identifying}.

State-of-the-art neural models for both tasks typically rely on lexical and syntactic features, such as part-of-speech tags~\cite{marcheggiani2017simple}, syntactic trees~\cite{roth2016neural, zhang2018graph, li2018unified}, and global decoding constraints~\cite{li2019dependency}. 
In particular, \citet{roth2016neural} argue that syntactic features are necessary to achieve competitive performance in dependency-based SRL. 
\citet{zhang2018graph} also showed that dependency tree features can further improve relation extraction performance.
Although syntactic features are no doubt helpful, a known challenge is that parsers are not available for every language, and even when available, they may not be sufficiently robust, especially for out-of-domain text, which may even hurt performance~\cite{he2017deep}.

Recently, the NLP community has seen excitement around neural models that make heavy use of pretraining based on language modeling~\cite{N18-1202,radford2018improving}.
The latest development is BERT \cite{devlin2018bert}, which has shown impressive gains in a wide variety of natural language tasks ranging from sentence classification to sequence labeling.
A natural question follows:\ can we leverage these pretrained models to further push the state of the art in relation extraction and semantic role labeling, without relying on lexical or syntactic features?
The answer is yes. 
We show that simple neural architectures built on top of BERT yields state-of-the-art performance on a variety of benchmark datasets for these two tasks.
The remainder of this paper describes our models and experimental results for relation extraction and semantic role labeling in turn.

\section{BERT for Relation Extraction}
\subsection{Model}

\begin{figure}[t]
	\centering
	\includegraphics[width=0.38\paperwidth]{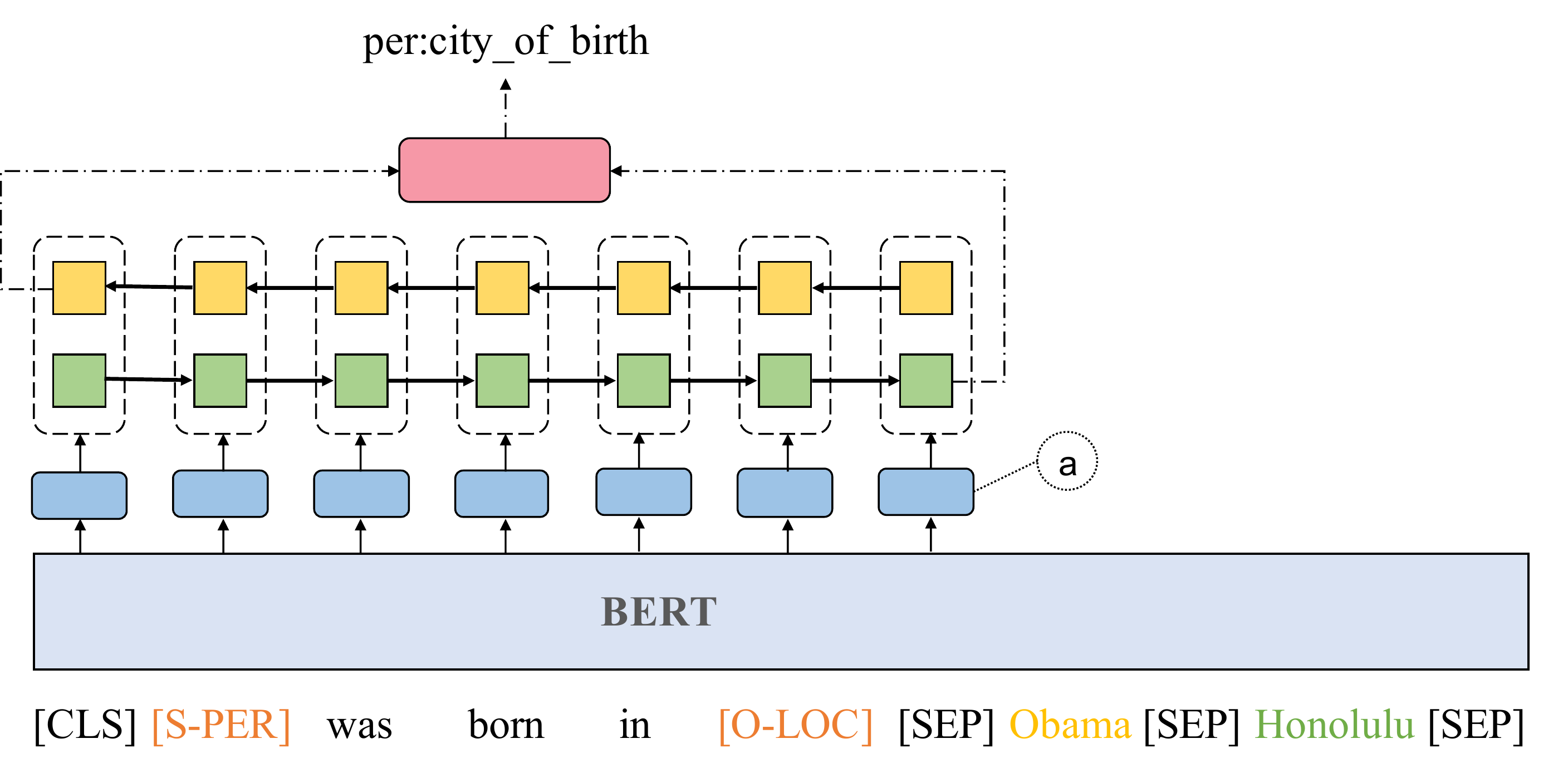}
	\caption{Architecture of our relation extraction model.
\textbf{(a)} denotes the concatenation of BERT contextual embedding and position embedding.
The final prediction is based on the concatenation of the final hidden state in each direction from the BiLSTM, fed through an MLP.}
	\label{fig:relation_arch}
\end{figure}

For relation extraction, the task is to predict the relation between two entities, given a sentence and two non-overlapping entity spans. 
In order to encode the sentence in an entity-aware manner, we propose the BERT-based model shown in Figure~\ref{fig:relation_arch}.
First, we construct the input sequence [[\textsc{cls}] sentence [\textsc{sep}] subject [\textsc{sep}] object [\textsc{sep}]]. 
To prevent overfitting, we replace the entity mentions in the sentence with masks, comprised of argument type~(subject or object) and entity type~(such as location and person), e.g., \textsc{Subj-Loc}, denoting that the subject entity is a location.

The input is then tokenized by the WordPiece tokenizer~\cite{sennrich2015neural} and fed into the BERT encoder. 
After obtaining the contextual representation, we discard the sequence after the first [\textsc{sep}] for the following operations.

We use $\mathcal{H} = [h_0, h_1, ..., h_n, h_{n+1}]$ to denote the BERT contextual representation for [[\textsc{cls}] sentence [\textsc{sep}]]. 
Note that $n$ can be different from the length of the sentence because the tokenizer might split words into sub-tokens. 
The subject entity span is denoted $\mathcal{H}_s = [h_{s_1}, h_{s_1 + 1}, ..., h_{s_2}]$ and similarly the object entity span is $\mathcal{H}_o = [h_{o_1}, h_{o_1+1}, ..., h_{o_2}]$. 
Following \citet{zhang2017position}, we define a position sequence relative to the subject entity span $[p_0^s, ..., p_{n+1}^s]$, where
\begin{equation}
	p_i^s =
		\begin{cases}
			i-s_1,      &  i < s_1\\
			0, &  s_1 < i < s_2 \\
			i-s_2, & i > s_2
		\end{cases}
\end{equation}
Here $s_1$ and $s_2$ are the starting and ending positions of the subject entity~(after tokenization), 
and $p_i^s \in \mathbb{Z}$ is the relative distance (in tokens) to the subject entity. 
A position sequence relative to the object $[p_0^o, ..., p_{n+1}^o]$ can be obtained in a similar way. 
To incorporate the position information into the model, the position sequences are converted into position embeddings, 
which are then concatenated to the contextual representation $\mathcal{H}$, followed by a one-layer BiLSTM. 
The final hidden states in each direction of the BiLSTM are used for prediction with a one-hidden-layer MLP.

\begin{table}[t]
	\centering
	\small
	\begin{tabular}{l@{\qquad}ccc}
		\toprule
		\textbf{Model}         & \textbf{P} & \textbf{R} & $\mathbf{F_1}$ \\ \midrule
		\citet{zhang2017position} &  65.7 & 64.5 & 65.1 \\ 
		\citet{zhang2018graph} & 69.9  & 63.33   &  66.4 \\
		\citet{wu2019simplifying} &  - & - & 67.0 \\ 
		\citet{alt2018improving} & 70.1 & 65.0 & 67.4 \\ \midrule
		BERT-LSTM-base & \textbf{73.3} & 63.10 & 67.8 \\  \midrule
		\citet{zhang2018graph} (ensemble) & 71.3 & \textbf{65.4} & \textbf{68.2} \\
		\bottomrule
	\end{tabular}
	\caption{Results on the \textsc{TACRED} test set.}
	\label{res:rel_tacred}
\end{table}

\subsection{Experiments}

We evaluate our model on the TAC Relation Extraction Dataset (TACRED)~\cite{zhang2017position}, a standard benchmark dataset for relation extraction.
In our experiments, the hidden sizes of the LSTM and MLP are 768 and 300, respectively, and the position embedding size is 20. The learning rate is $5 \times 10^{-5}$.  
The BERT base-cased model is used in our experiments. 
Embeddings for the masks (e.g., \textsc{Subj-Loc}) are randomly initialized and fine-tuned during the training process, as well as the position embeddings. 

Results on the TACRED test set are shown in Table~\ref{res:rel_tacred}. 
Our model outperforms the works of \citet{zhang2018graph} and \citet{wu2019simplifying}, which use GCNs~\cite{kipf2016semi} and variants to encode syntactic tree information as external features. 
\citet{alt2018improving} leverage the pretrained language model GPT~\cite{radford2018improving} and achieves better recall than our system. 
In terms of $F_1$, our system obtains the best known score among {\it individual} models, but our score is still below that of the interpolation model of~\citet{zhang2018graph} because of lower recall. 

\section{BERT for Semantic Role Labeling}

\subsection{Model}

The standard formulation of semantic role labeling decomposes into four subtasks:\ predicate detection, predicate sense disambiguation, argument identification, and argument classification. 
There are two representations for argument annotation:\ span-based and dependency-based. 
Semantic banks such as PropBank usually represent arguments as syntactic constituents~(spans), whereas the CoNLL 2008 and 2009 shared tasks propose dependency-based SRL, where the goal is to identify the syntactic heads of arguments rather than the entire span. 
Here, we follow~\citet{li2019dependency} to unify these two annotation schemes into one framework, without any declarative constraints for decoding. 
For several SRL benchmarks, such as CoNLL 2005, 2009, and 2012, the predicate is given during both training and testing. 
Thus, in this paper, we only discuss predicate disambiguation and argument identification and classification.

\smallskip \noindent {\bf Predicate sense disambiguation.}
The predicate disambiguation task is to identify the correct meaning of a predicate in a given context. 
As an example, for the sentence ``Barack Obama went to Paris'', the predicate \textit{went} has sense ``motion'' and has sense label \textit{01}.

We formulate this task as sequence labeling.
The input sentence is fed into the WordPiece tokenizer, which splits some words into sub-tokens.
The predicate token is tagged with the sense label.
Following the original BERT paper, two labels are used for the remaining tokens:\ `O' for the first (sub-)token of any word and `X' for any remaining fragments. 
We feed the sequences into the BERT encoder to obtain the contextual representation $\mathcal{H}$. 
A ``predicate indicator'' embedding is then concatenated to the contextual representation to distinguish the predicate tokens from non-predicate ones. 
The final prediction is made using a one-hidden-layer MLP over the label set.

\begin{figure}[t]
	\centering
	\includegraphics[width=0.33\paperwidth]{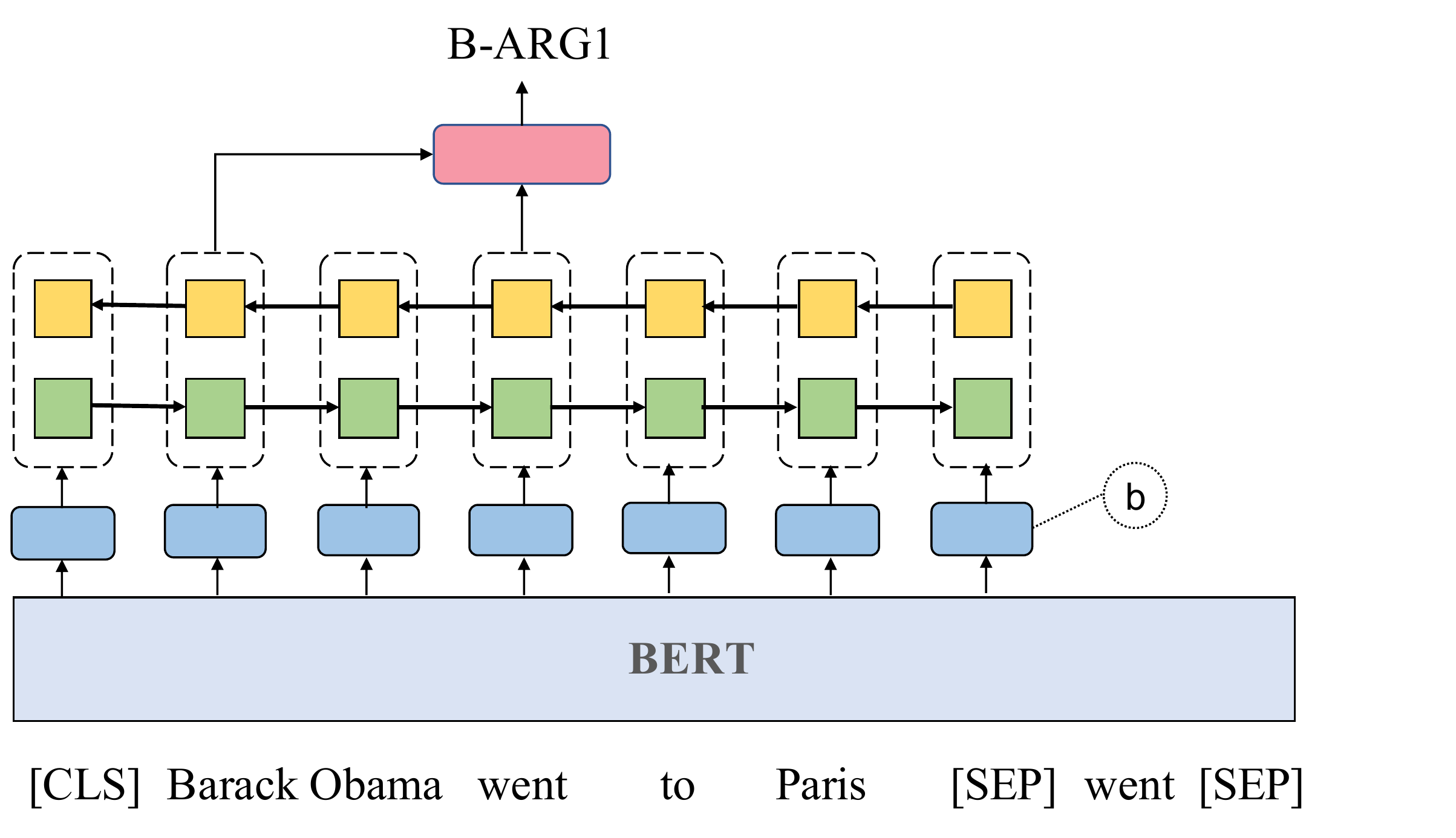}
	\caption{Architecture of our predicate identification and classification model, at the point where the model is making a prediction for the token ``Barack''.
		\textbf{(b)} denotes the concatenation of BERT contextual embedding and predicate indicator embedding. The final prediction is based on the concatenation of the hidden state of the predicate (``went'') and the hidden state of the current token, fed through an MLP.}
	\label{fig:srl_arch}
\end{figure}

\smallskip \noindent {\bf Argument identification and classification.}
This task is to detect the argument spans or argument syntactic heads and assign them the correct semantic role labels. 
In the above example, ``Barack Obama'' is the \textsc{Arg1} of the predicate \textit{went}, meaning the entity in motion.

\begin{table}[t]
	\centering
	\small
	\begin{tabular}{l@{\qquad}ccc}
		\toprule
		\textbf{Model}         & \textbf{Dev} & \textbf{Test} & \textbf{Brown} \\ \midrule
		\citet{shi2017joint} &  - & 93.43 & 82.36 \\
		\citet{roth2016neural} & 94.77 & 95.47 & - \\
		\citet{he2018syntax} & 95.01 & 95.58 & - \\ \midrule
		BERT-base & \textbf{96.32 }& \textbf{96.88} &  \textbf{90.63}  \\ 
		\bottomrule
	\end{tabular}
	\caption{Predicate disambiguation accuracy on the CoNLL 2009 dataset.}
	\label{res:preddis}
\end{table}

\begin{table}[t]
	\centering
	\small
	\begin{tabular}{l ccc}
		\toprule
		\textbf{Model}      &  \textbf{Dev} & \textbf{Test} & \textbf{Brown} \\ \midrule
		\citet{marcheggiani2017encoding} &  83.3 & - & - \\ 
		\citet{he2018syntax} & 84.2 & - & - \\
		\citet{shi2017joint} & 85.6 & 87.1  & 77.4  \\
		\midrule 
		BERT-LSTM-base & 88.7 & 89.8 & 82.7\\
		BERT-LSTM-large & \textbf{89.3} & \textbf{90.3} & \textbf{83.5} \\
		\bottomrule
	\end{tabular}
	\caption{Comparison of $F_1$ scores for argument identification and classification on the CoNLL 2009 dataset, excluding predicate sense disambiguation.}
	\label{res:conll09_exclude}
\end{table}

Formally, our task is to predict a sequence $\boldsymbol{z}$ given a sentence--predicate pair ($\mathcal{X}$, $v$) as input, where the label set draws from the cross of the standard BIO tagging scheme and the arguments of the predicate (e.g., \textsc{B-Arg1}).
The model architecture is illustrated in Figure~\ref{fig:srl_arch}, at the point in the inference process where it is outputting a tag for the token ``Barack''.
In order to encode the sentence in a predicate-aware manner, we design the input as [[\textsc{cls}] sentence [\textsc{sep}] predicate [\textsc{sep}]], allowing the representation of the predicate to interact with the entire sentence via appropriate attention mechanisms.
The input sequence as described above is fed into the BERT encoder. 
The contextual representation of the sentence~([\textsc{cls}] sentence [\textsc{sep}]) from BERT is then concatenated to predicate indicator embeddings, followed by a one-layer BiLSTM to obtain hidden states $\mathcal{G} = [g_1, g_2, ..., g_n]$. 
For the final prediction on each token $g_{i}$, the hidden state of predicate $g_{p}$ is concatenated to the hidden state of the token $g_{i}$, and then fed into a one-hidden-layer MLP classifier over the label set.

\begin{table*}[t]
	\centering
	\small
	\begin{tabular}{ll ccc ccc}
		\toprule
		&  & \multicolumn{3}{c}{CoNLL 09 (In-domain)} &  \multicolumn{3}{c}{Out-of-domain (Brown)} \\ 
		\cmidrule(lr){3-5} \cmidrule(lr){6-8}
		&Model  & P & R & $F_1$ & P & R & $F_1$ \\
		\midrule
		\multirow{5}{*}{Single} & \citet{he2018syntax}  & 89.7 & 89.3 & 89.5 & 81.9 & 76.9 & 79.3 \\
		&   \citet{li2018unified} & 90.3 & 89.3 & 89.8 & 80.6 & 79.0 & 79.8 \\
		&	\citet{li2019dependency}  & 89.6 & 91.2 & 90.4 & 81.7 & 81.4 & 81.5 \\  \cmidrule{2-8}
		& BERT-LSTM-base & 92.1 & 91.9 & 92.0 & 85.6 & 84.7 & 85.1 \\
		& BERT-LSTM-large & \textbf{92.4} & \textbf{92.3} & \textbf{92.4} & \textbf{85.7} & \textbf{85.8 } & \textbf{85.7} \\
		\midrule
		
		\multirow{2}{*}{Ensemble} &	\citet{roth2016neural}  &90.3 & 85.7 & 87.9 & 79.7 & 73.6 & 76.5 \\
		&	\citet{marcheggiani2017encoding}  & 90.5 & 87.7 & 89.1 & 80.8 & 77.1 & 78.9 \\ 
		\bottomrule
	\end{tabular}
	\caption{Performance comparison on dependency-based SRL.\label{res:conll09}}
\end{table*}

\begin{table*}[t]
	\centering
	\small
	\begin{tabular}{l ccc ccc ccc}
		\toprule
		& \multicolumn{3}{c}{CoNLL 05 (In-domain)} &  \multicolumn{3}{c}{Out-of-domain (Brown)} & \multicolumn{3}{c}{CoNLL 12 (In-domain)}  \\ 
		\cmidrule(lr){2-4} \cmidrule(lr){5-7} \cmidrule{8-10}
		Model  & P & R & $F_1$ & P & R & $F_1$ & P & R & $F_1$ \\
		\midrule
		\citet{strubell2018linguistically} & 86.0 & 86.0 & 86.0 & 76.7 & 76.4 & 76.5  & - & - & - \\
		\citet{he2018jointly} & - & - & 87.4 & - & - & 80.4  & - & - & 85.5 \\
		\citet{ouchi2018span} & 88.2 & 87.0 & 87.6 & 79.9 & 77.5 & 78.7  &87.1 & 85.3 & 86.2 \\  
		\citet{li2019dependency}  & 87.9 & 87.5 & 87.7 & 80.6 & 80.4 & 80.5  & 85.7 & 86.3 & 86.0\\  \cmidrule{1-10}
		BERT-LSTM-base  & 87.8 & 88.4 & 88.1 & 80.7 & 81.2 & 80.9  & 85.7 & 86.7 & 86.2\\
		BERT-LSTM-large  & 88.6 & \textbf{89.0} & \textbf{88.8} & \textbf{81.9} &\textbf{82.1}&\textbf{82.0 } &  85.9 & \textbf{87.0} &  86.5\\
		\midrule
		\citet{ouchi2018span}~(ensemble)  &\textbf{89.2 }& 87.9 & 88.5 & 81.0 & 78.4 & 79.6 &  \textbf{88.5} & 85.5& \textbf{87.0}\\
		\bottomrule
	\end{tabular}
	\caption{Performance comparison on span-based SRL.}
	\label{res:conll12}
	\vspace{-1mm}
\end{table*}

\subsection{Experimental Setup}

We conduct experiments on two SRL tasks:\ span-based and dependency-based. 
For span-based SRL, the CoNLL 2005~\cite{carreras2004introduction} and 2012~\cite{pradhan2013towards} datasets are used. 
For dependency-based SRL, the CoNLL 2009~\cite{hajivc2009conll} dataset is used. 
We follow standard splits for the training, development, and test sets.

In our experiments, the hidden sizes of the LSTM and MLP are 768 and 300, respectively, and the predicate indicator embedding size is 10.
The learning rate is $5 \times 10^{-5}$.  
BERT base-cased and large-cased models are used in our experiments. 
The position embeddings are randomly initialized and fine-tuned during the training process.

\subsection{Dependency-Based SRL Results}

\noindent \textbf{Predicate sense disambiguation.} 
The predicate sense disambiguation subtask applies only to the CoNLL 2009 benchmark.
In this line of research on dependency-based SRL, previous papers seldom report the accuracy of predicate disambiguation separately~(results are often mixed with argument identification and classification), causing difficulty in determining the source of gains. 
Here, we report predicate disambiguation accuracy in Table~\ref{res:preddis} for the development set, test set, and the out-of-domain test set~(Brown). 
The state-of-the-art model~\cite{he2018syntax} is based on a BiLSTM and linguistic features such as POS tag embeddings and lemma embeddings. 
Instead of using linguistic features, our simple MLP model achieves better accuracy with the help of powerful contextual embeddings. 
These predicate sense disambiguation results are used in the dependency-based SRL end-to-end evaluation.

\smallskip \noindent \textbf{Argument identification and classification.} 
We provide SRL performance excluding predicate sense disambiguation to validate the source of improvements:\ results are shown in Table~\ref{res:conll09_exclude}.
Figures from some systems are missing because they only report end-to-end results.

Our end-to-end results are shown in Table~\ref{res:conll09}.
We see that the BERT-LSTM-large model~(using the predicate sense disambiguation results from above) yields large $F_1$ score improvements over the existing state of the art~\cite{li2019dependency}, and beats existing ensemble models as well.
This is achieved without using any linguistic features and declarative decoding constraints.

\subsection{Span-Based SRL Results}

Our span-based SRL results are shown in Table~\ref{res:conll12}. 
We see that the BERT-LSTM-large model achieves the state-of-the-art $F_1$ score among single models and outperforms the \citet{ouchi2018span} ensemble model on the CoNLL 2005 in-domain and out-of-domain tests. 
However, it falls short on the CoNLL 2012 benchmark because the model of \citet{ouchi2018span} obtains very high precision. 
They are able to achieve this with a more complex decoding layer, with human-designed constraints such as the ``Overlap Constraint'' and ``Number Constraint''.

\section{Conclusions}

Based on this preliminary study, we show that BERT can be adapted to relation extraction and semantic role labeling without syntactic features and human-designed constraints. 
While we concede that our model is quite simple, we argue this is a feature, as the power of BERT is able to simplify neural architectures tailored to specific tasks.
Nevertheless, these results provide strong baselines and foundations for future research.
Many natural follow-up questions emerge:\
Can syntactic features be re-introduced to further improve results?
Can multitask learning be used to simultaneously benefit relation extraction and semantic role labeling?
We are actively working on answering these and additional questions.

\section*{Acknowledgments}

This research was supported by the Natural Sciences and Engineering Research Council (NSERC) of Canada. 


\begin{thebibliography}{25}
\expandafter\ifx\csname natexlab\endcsname\relax\def\natexlab#1{#1}\fi

\bibitem[{Alt et~al.(2019)Alt, H{\"u}bner, and Hennig}]{alt2018improving}
Christoph Alt, Marc H{\"u}bner, and Leonhard Hennig. 2019.
\newblock Improving relation extraction by pre-trained language
  representations.
\newblock In \emph{AKBC}.

\bibitem[{Carreras and M{\`a}rquez(2004)}]{carreras2004introduction}
Xavier Carreras and Llu{\'\i}s M{\`a}rquez. 2004.
\newblock Introduction to the {CoNLL}-2004 shared task: Semantic role labeling.
\newblock In \emph{Proceedings of the Eighth Conference on Computational
  Natural Language Learning (CoNLL-2004) at HLT-NAACL 2004}.

\bibitem[{Devlin et~al.(2018)Devlin, Chang, Lee, and
  Toutanova}]{devlin2018bert}
Jacob Devlin, Ming-Wei Chang, Kenton Lee, and Kristina Toutanova. 2018.
\newblock {BERT}: Pre-training of deep bidirectional transformers for language
  understanding.
\newblock \emph{arXiv:1810.04805}.

\bibitem[{Fader et~al.(2011)Fader, Soderland, and
  Etzioni}]{fader2011identifying}
Anthony Fader, Stephen Soderland, and Oren Etzioni. 2011.
\newblock Identifying relations for open information extraction.
\newblock In \emph{Proceedings of the 2011 Conference on Empirical Methods in
  Natural Language Processing}, pages 1535--1545.

\bibitem[{Haji{\v{c}} et~al.(2009)Haji{\v{c}}, Ciaramita, Johansson, Kawahara,
  Mart{\'\i}, M{\`a}rquez, Meyers, Nivre, Pad{\'o}, {\v{S}}t{\v{e}}p{\'a}nek
  et~al.}]{hajivc2009conll}
Jan Haji{\v{c}}, Massimiliano Ciaramita, Richard Johansson, Daisuke Kawahara,
  Maria~Ant{\`o}nia Mart{\'\i}, Llu{\'\i}s M{\`a}rquez, Adam Meyers, Joakim
  Nivre, Sebastian Pad{\'o}, Jan {\v{S}}t{\v{e}}p{\'a}nek, et~al. 2009.
\newblock The {CoNLL-2009} shared task: Syntactic and semantic dependencies in
  multiple languages.
\newblock In \emph{Proceedings of the Thirteenth Conference on Computational
  Natural Language Learning:\ Shared Task}, pages 1--18.

\bibitem[{He et~al.(2018{\natexlab{a}})He, Lee, Levy, and
  Zettlemoyer}]{he2018jointly}
Luheng He, Kenton Lee, Omer Levy, and Luke Zettlemoyer. 2018{\natexlab{a}}.
\newblock Jointly predicting predicates and arguments in neural semantic role
  labeling.
\newblock In \emph{Proceedings of the 56th Annual Meeting of the Association
  for Computational Linguistics (Volume 2:\ Short Papers)}, pages 364--369.

\bibitem[{He et~al.(2017)He, Lee, Lewis, and Zettlemoyer}]{he2017deep}
Luheng He, Kenton Lee, Mike Lewis, and Luke Zettlemoyer. 2017.
\newblock Deep semantic role labeling: What works and what{'}s next.
\newblock In \emph{Proceedings of the 55th Annual Meeting of the Association
  for Computational Linguistics (Volume 1:\ Long Papers)}, pages 473--483.

\bibitem[{He et~al.(2018{\natexlab{b}})He, Li, Zhao, and Bai}]{he2018syntax}
Shexia He, Zuchao Li, Hai Zhao, and Hongxiao Bai. 2018{\natexlab{b}}.
\newblock Syntax for semantic role labeling, to be, or not to be.
\newblock In \emph{Proceedings of the 56th Annual Meeting of the Association
  for Computational Linguistics (Volume 1: Long Papers)}, pages 2061--2071.

\bibitem[{Kipf and Welling(2016)}]{kipf2016semi}
Thomas~N. Kipf and Max Welling. 2016.
\newblock Semi-supervised classification with graph convolutional networks.
\newblock In \emph{Proceedings of the 5th International Conference on Learning
  Representations}.

\bibitem[{Li et~al.(2018)Li, He, Cai, Zhang, Zhao, Liu, Li, and
  Si}]{li2018unified}
Zuchao Li, Shexia He, Jiaxun Cai, Zhuosheng Zhang, Hai Zhao, Gongshen Liu,
  Linlin Li, and Luo Si. 2018.
\newblock A unified syntax-aware framework for semantic role labeling.
\newblock In \emph{Proceedings of the 2018 Conference on Empirical Methods in
  Natural Language Processing}, pages 2401--2411.

\bibitem[{Li et~al.(2019)Li, He, Zhao, Zhang, Zhang, Zhou, and
  Zhou}]{li2019dependency}
Zuchao Li, Shexia He, Hai Zhao, Yiqing Zhang, Zhuosheng Zhang, Xi~Zhou, and
  Xiang Zhou. 2019.
\newblock Dependency or span, end-to-end uniform semantic role labeling.
\newblock In \emph{Proceedings of the 33rd AAAI Conference on Artificial
  Intelligence}.

\bibitem[{Marcheggiani et~al.(2017)Marcheggiani, Frolov, and
  Titov}]{marcheggiani2017simple}
Diego Marcheggiani, Anton Frolov, and Ivan Titov. 2017.
\newblock A simple and accurate syntax-agnostic neural model for
  dependency-based semantic role labeling.
\newblock In \emph{Proceedings of the 21st Conference on Computational Natural
  Language Learning (CoNLL 2017)}, pages 411--420.

\bibitem[{Marcheggiani and Titov(2017)}]{marcheggiani2017encoding}
Diego Marcheggiani and Ivan Titov. 2017.
\newblock Encoding sentences with graph convolutional networks for semantic
  role labeling.
\newblock In \emph{Proceedings of the 2017 Conference on Empirical Methods in
  Natural Language Processing}, pages 1506--1515.

\bibitem[{Ouchi et~al.(2018)Ouchi, Shindo, and Matsumoto}]{ouchi2018span}
Hiroki Ouchi, Hiroyuki Shindo, and Yuji Matsumoto. 2018.
\newblock A span selection model for semantic role labeling.
\newblock In \emph{Proceedings of the 2018 Conference on Empirical Methods in
  Natural Language Processing}, pages 1630--1642.

\bibitem[{Peters et~al.(2018)Peters, Neumann, Iyyer, Gardner, Clark, Lee, and
  Zettlemoyer}]{N18-1202}
Matthew Peters, Mark Neumann, Mohit Iyyer, Matt Gardner, Christopher Clark,
  Kenton Lee, and Luke Zettlemoyer. 2018.
\newblock Deep contextualized word representations.
\newblock In \emph{Proceedings of the 2018 Conference of the North American
  Chapter of the Association for Computational Linguistics:\ Human Language
  Technologies, Volume 1 (Long Papers)}, pages 2227--2237.

\bibitem[{Pradhan et~al.(2013)Pradhan, Moschitti, Xue, Ng, Bj{\"o}rkelund,
  Uryupina, Zhang, and Zhong}]{pradhan2013towards}
Sameer Pradhan, Alessandro Moschitti, Nianwen Xue, Hwee~Tou Ng, Anders
  Bj{\"o}rkelund, Olga Uryupina, Yuchen Zhang, and Zhi Zhong. 2013.
\newblock Towards robust linguistic analysis using {OntoNotes}.
\newblock In \emph{Proceedings of the Seventeenth Conference on Computational
  Natural Language Learning}, pages 143--152.

\bibitem[{Radford et~al.(2018)Radford, Narasimhan, Salimans, and
  Sutskever}]{radford2018improving}
Alec Radford, Karthik Narasimhan, Tim Salimans, and Ilya Sutskever. 2018.
\newblock Improving language understanding by generative pre-training.

\bibitem[{Roth and Lapata(2016)}]{roth2016neural}
Michael Roth and Mirella Lapata. 2016.
\newblock Neural semantic role labeling with dependency path embeddings.
\newblock In \emph{Proceedings of the 54th Annual Meeting of the Association
  for Computational Linguistics (Volume 1: Long Papers)}, pages 1192--1202.

\bibitem[{Sennrich et~al.(2016)Sennrich, Haddow, and
  Birch}]{sennrich2015neural}
Rico Sennrich, Barry Haddow, and Alexandra Birch. 2016.
\newblock Neural machine translation of rare words with subword units.
\newblock In \emph{Proceedings of the 54th Annual Meeting of the Association
  for Computational Linguistics (Volume 1: Long Papers)}, pages 1715--1725.

\bibitem[{Shen and Lapata(2007)}]{shen2007using}
Dan Shen and Mirella Lapata. 2007.
\newblock Using semantic roles to improve question answering.
\newblock In \emph{Proceedings of the 2007 Joint Conference on Empirical
  Methods in Natural Language Processing and Computational Natural Language
  Learning (EMNLP-CoNLL)}, pages 12--21.

\bibitem[{Shi and Zhang(2017)}]{shi2017joint}
Peng Shi and Yue Zhang. 2017.
\newblock Joint bi-affine parsing and semantic role labeling.
\newblock In \emph{Proceedings of the 2017 International Conference on Asian
  Language Processing (IALP)}, pages 338--341.

\bibitem[{Strubell et~al.(2018)Strubell, Verga, Andor, Weiss, and
  McCallum}]{strubell2018linguistically}
Emma Strubell, Patrick Verga, Daniel Andor, David Weiss, and Andrew McCallum.
  2018.
\newblock Linguistically-informed self-attention for semantic role labeling.
\newblock In \emph{Proceedings of the 2018 Conference on Empirical Methods in
  Natural Language Processing}, pages 5027--5038.

\bibitem[{Wu et~al.(2019)Wu, Zhang, Souza~Jr, Fifty, Yu, and
  Weinberger}]{wu2019simplifying}
Felix Wu, Tianyi Zhang, Amauri Holanda~de Souza~Jr, Christopher Fifty, Tao Yu,
  and Kilian~Q. Weinberger. 2019.
\newblock Simplifying graph convolutional networks.
\newblock \emph{arXiv:1902.07153}.

\bibitem[{Zhang et~al.(2018)Zhang, Qi, and Manning}]{zhang2018graph}
Yuhao Zhang, Peng Qi, and Christopher~D. Manning. 2018.
\newblock Graph convolution over pruned dependency trees improves relation
  extraction.
\newblock In \emph{Proceedings of the 2018 Conference on Empirical Methods in
  Natural Language Processing}, pages 2205--2215.

\bibitem[{Zhang et~al.(2017)Zhang, Zhong, Chen, Angeli, and
  Manning}]{zhang2017position}
Yuhao Zhang, Victor Zhong, Danqi Chen, Gabor Angeli, and Christopher~D.
  Manning. 2017.
\newblock Position-aware attention and supervised data improve slot filling.
\newblock In \emph{Proceedings of the 2017 Conference on Empirical Methods in
  Natural Language Processing}, pages 35--45.

\end{thebibliography}

\end{document}